\documentclass{article}

\usepackage{microtype}
\usepackage{graphicx}
\usepackage{booktabs} 

\usepackage{subcaption}
\usepackage[utf8]{inputenc} 
\usepackage[T1]{fontenc}    
\usepackage{url}            
\usepackage{amsfonts}       
\usepackage{amsmath}	

\graphicspath{{./graphics/}{./graphics/wild/}{./graphics/mnist/}{./graphics/micro/}{./graphics/food/}{./graphics/fashion/}{./graphics/missingrate/}}

\def\z{{\bf z}}
\def\x{{\bf x}}

\def\I{{\bf I}}

\def\epsilo{\boldsymbol\epsilon}

\def\0{{\bf 0}}
\def\xb{{\bf x}}

\def\Lambdab{\text{\boldmath $\Lambda$}}
\def\Ub{{\bf U}}

\def\vb{{\bf v}}

\def\ub{{\bf u}}

\def\rb{{\bf r}}
\def\Qb{{\bf Q}}

\def\ab{{\bf a}}
\def\Ab{{\bf A}}
\def\Cb{{\bf C}}

\def\Ib{{\bf I}}
\def\Mb{{\bf M}}
\def\Xb{{\bf X}}
\def\Ib{{\bf I}}
\def\0b{{\bf 0}}
\def\hb{{\bf h}}
\def\sb{{\bf s}}

\def\z{{\bf z}}
\def\x{{\bf x}}

\def\I{{\bf I}}

\def\epsilo{\boldsymbol\epsilon}

\def\0{{\bf 0}}
\usepackage{hyperref}

\usepackage[accepted]{icml2019}

\icmltitlerunning{Phase transition in PCA with missing data: Reduced signal-to-noise ratio, not sample size!}

\begin{document}

\twocolumn[
\icmltitle{Phase transition in PCA with missing data:\\
		Reduced signal-to-noise ratio, not sample size!}



\icmlsetsymbol{equal}{*}

\begin{icmlauthorlist}
\icmlauthor{Niels Bruun Ipsen}{dtu}
\icmlauthor{Lars Kai Hansen}{dtu}
\end{icmlauthorlist}

\icmlaffiliation{dtu}{Department of Applied Mathematics and Computer Science, Technical University of Denmark, Denmark}

\icmlcorrespondingauthor{Niels Bruun Ipsen}{nbip@dtu.dk}
\icmlcorrespondingauthor{Lars Kai Hansen}{lkai@dtu.dk}

\icmlkeywords{Machine Learning, ICML}

\vskip 0.3in
]



\printAffiliationsAndNotice{}  

\begin{abstract}
How does missing data affect our ability to learn signal structures? It has been shown that learning signal structure in terms of principal components is dependent on the ratio of sample size and dimensionality and that a critical number of observations is needed before learning starts (Biehl and Mietzner, 1993). Here we generalize this analysis to include missing data. Probabilistic principal component analysis is regularly used for estimating signal structures in datasets with missing data.
Our analytic result suggests that the effect of  missing data  is to effectively reduce signal-to-noise ratio rather than - as generally believed - to reduce sample size. The theory predicts a phase transition in the learning curves and this is indeed found both in simulation data and in real datasets.
\end{abstract}

\section{Introduction}
Principal component analysis (PCA) is a  widely used tool for exploratory data analysis which originates with Karl Pearson \yrcite{pearson1901liii} who described a method for obtaining the "best fit" of a plane or a line to a system of points, by minimizing the sum of squared orthogonal distances from the points to the line/plane. PCA  has been rediscovered many times, for example by Hotelling \yrcite{hotelling1933analysis}. In many research areas variants of PCA are known under specific names, viz., the Kosambi-Karhunen-Lo\`{e}ve transform in signal processing, the Hotelling transform in multivariate quality control and Latent Semantic Analysis in natural language processing.
A well known and illustrative application of PCA is the eigenfaces investigation by Turk and Pentland \yrcite{turk1991eigenfaces}.\\
The probabilistic version of PCA was proposed independently by Roweis \yrcite{roweis1998algorithms} and Tipping and Bishop \yrcite{tipping1999probabilistic} and this formulation allows for estimating principal components in the presence of missing data via the EM algorithm \cite{dempster1977maximum}.
 Numerous modern PCA applications are reviewed in the recent volume \cite{sanguansat2012principal} and new variations on the classical scheme keep developing, see e.g., a recent  `supervised' principal component analysis for microbiology \cite{2017Vehtari}.

Explorative data analysis is particularly relevant in high dimensional and correlated measurements such as images, times series, frequency spectra or text and where the data generating process yields relatively `clean'  measurements, i.e., the variance of the signal of interest $\sigma^2_{\rm signal}$ dominates the variance of the measurement noise $\sigma^2_{\rm noise}$ leaving a signal-to-noise variance ratio bigger than one: $S=\sigma^2_{\rm signal}/\sigma^2_{\rm noise} > 1$. The structure of such a clean signal can be learned from data by PCA: Let the measured signal be a $D-$dimensional vector $\xb$ and let the training sample consist of $N$ such measurements $\mathcal{X} = \{\xb_1,\xb_2,...\xb_N\} $. Assuming that a non-informative mean has been subtracted from the data we can infer  the pattern of covariance in measurement space $\widehat{\ab}$  that maximally `explains'  variance, as $\widehat{\ab} = {\arg \max}_{||\ab|| = 1 } \sum_{n=1}^N (\ab^{T}\cdot\xb_n)^2$. Empirically, it is found that for sample sizes large enough, the estimate $\widehat{\ab}$ stabilizes. In simulation studies where the ground truth direction $\ab_0$ is known we can use the quantity $R^2 = (\widehat{\ab}^{\top}\cdot\ab_0)^2 $ to measure the alignment between estimated and ground truth directions and there we find that as $N \rightarrow \infty$ perfect alignment $R^2 \rightarrow 1 $ is obtained under weak conditions.

By the wide applicability, PCA has also gained significant theoretical interest and a rather complete analytical understanding has been obtained in the big data limit $D,N\rightarrow \infty $  \citep{biehl1993statistical,hoyle2007statistical}. In fact in this limit a remarkable universality is found. If we define the ratio of samples to dimensionality $\alpha = N/D$, the analytical result
\begin{equation}
\langle  R^2 \rangle_{\mathcal{X}} =
    \begin{cases}
     0	&	\alpha S^{2} < 1,  \\
     \frac{\alpha S^2 - 1}{S + \alpha S^2}	&	\alpha S^{2} \geq 1
	\end{cases}
	\label{eq:one}
\end{equation}
resembles a `phase transition' at the critical value of the effective sample size: $\alpha_{\rm critical}= S^{-2}$. The relation between signal-to-noise ratio and effective sample size is highly non-linear:  If PCA is presented with too small an effective sample size, we essentially learn nothing $\langle  R^2 \rangle_{\mathcal{X}} = 0$. Important for applications PCA is quite robust to sample size in the large sample size limit:  $\langle  R^2\rangle_{\mathcal{X}} = 1 - \frac{S+1}{S^2}\frac{1}{\alpha} + o\left(\frac{1}{\alpha^2}\right)$. The theoretical results are accurate already for modest sample sizes $N \sim 10^2-10^3$.

\section{Learning from incomplete data}
While conventional PCA can be said to be well-understood, an additional complication occurs often in practical applications. Namely, that the sampling process is imperfect leading to so-called `missing data', see e.g., \cite{van2014missing} for an example involving explorative analysis of questionnaires. How does missing data affect our ability to learn signal structures by PCA?

Conventional wisdom is that missing data leads to effectively smaller sample size, viz., the following statement in the highly cited review by Schafer and Graham \yrcite{schafer2002missing}: \emph{`In missing data problems the sample may have to be larger than usual, because missing values effectively reduce the sample size'}. While it is true that the problems of missing data can be mitigated by increasing  sample size, we  here  show that missing data is more accurately described as effectively reducing the signal-to-noise ratio.

 A simple approach to handling missing data is to `impute' the missing values, however, the specific imputation method  may significantly impact the results \cite{dray2015principal}.
  A probabilistic approach can formulate the inference problem based on the observed values only, assuming that the missing mechanism is 'missing at random', MAR, a weaker assumption than 'missing completely at random', MCAR \cite{little2014statistical}. For example, so-called probabilistic PCA  allows inference of principal components in the face of missing data, without imputation, when assuming the missing mechanism is MAR.
\section{Extending replica analysis to handle missing data}
While there are numerous empirical studies of imputation schemes and their performance, there has so far not been reported attempts to generalize Eq.\ (\ref{eq:one}) to missing data.
We here present a theoretical analysis expanding the already mentioned analytical results for PCA learning to include the effects of missing data.
While intuitively one would expect that missing data could take the form of an effective reduction of the sample size $N$,  we find analytically that when data is 'missing completely at random' \cite{little2014statistical} at a rate of $m$, the role of missing data is instead to reduce the effective signal-to-noise ratio. 
\begin{figure}[ht]
  \centering
  \includegraphics[width=\linewidth]{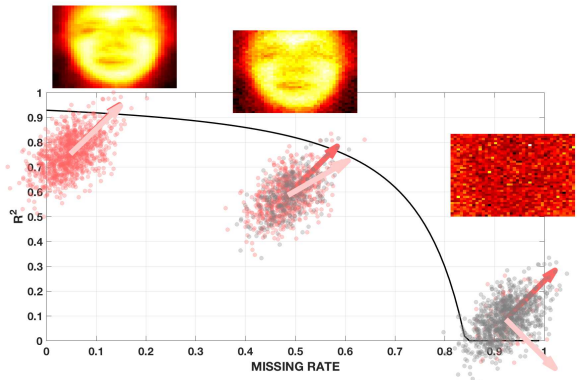}
  \caption{Conceptually, missing data impacts the cosine similarity, $R^2,$ between the true signal direction and the signal direction as estimated by probabilistic PCA. As the missing rate increases the estimated principal direction will deviate more and more from the true signal direction and $R^2$ will decrease according to Eq.\ (\ref{eq:two}). In a high dimensional dataspace, the estimated principal direction will on average be orthogonal to the true signal direction when the missing rate is higher than some critical value.
  Conventional wisdom is that the impact of missing data is similar to a reduced sample size. We argue that a more accurate description is that the impact of missing data is to reduce the signal-to-noise ratio.}
  \label{fig:fig1}
\end{figure}
%

While the details are quite involved and deferred to the appendix, a surprisingly simple result is found, namely a straightforward generalization of Eq.\ (\ref{eq:one}) with $S$ replaced by an effective signal-to-noise ratio $S(m) = (1-m)S$,
\begin{equation}
\langle  R^2 \rangle_{\mathcal{X}} = 	%
    \begin{cases}\vspace*{-1mm}
          0	&	\alpha S(m)^{2} <1,  \\
           \frac{\alpha S(m)^2 - 1}{S(m) + \alpha S(m)^2}	&	\alpha S(m)^{2} \geq 1.
	\end{cases}
    \label{eq:two}
\end{equation}

\begin{figure}[ht]
\vskip 0.2in
    \begin{center}
    \begin{subfigure}[b]{0.45\linewidth}
        \includegraphics[width=\columnwidth]{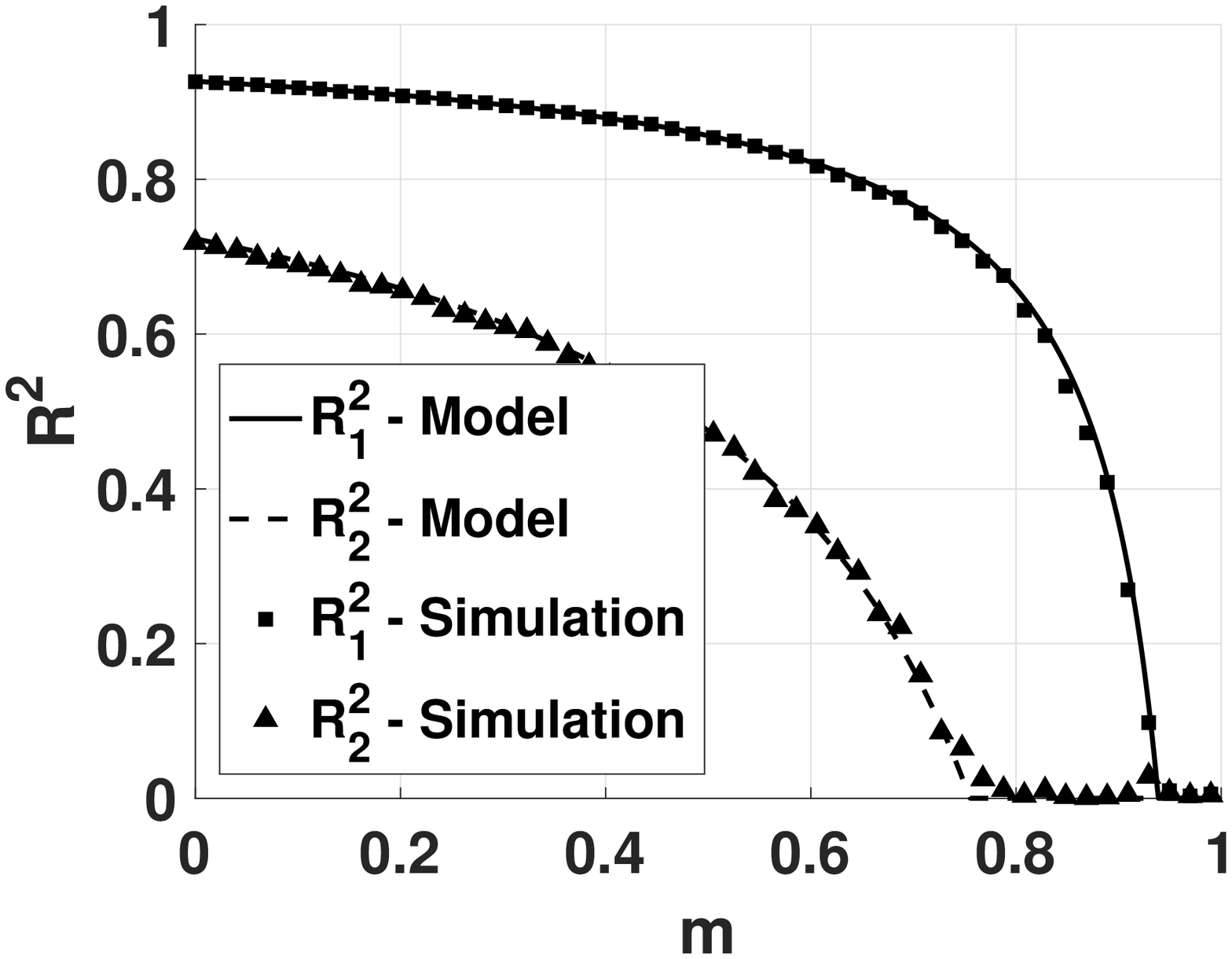}
        \caption{ }
        \label{fig:one}
    \end{subfigure}
    ~ 
    \begin{subfigure}[b]{0.45\linewidth}
        \includegraphics[width=\textwidth]{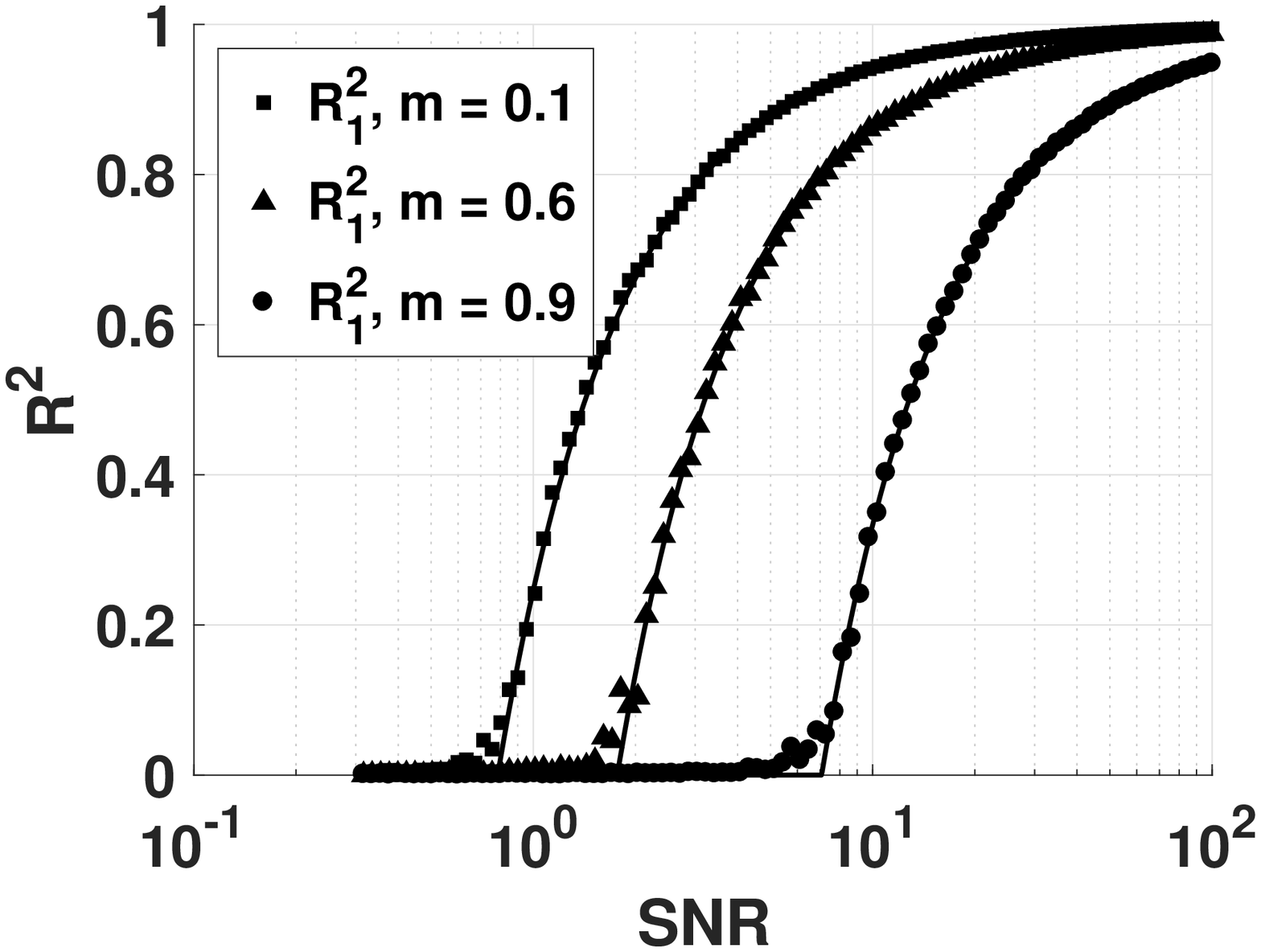}
        \caption{ }
        \label{fig:two}
    \end{subfigure}
        \begin{subfigure}[b]{0.45\linewidth}
        \includegraphics[width=\textwidth]{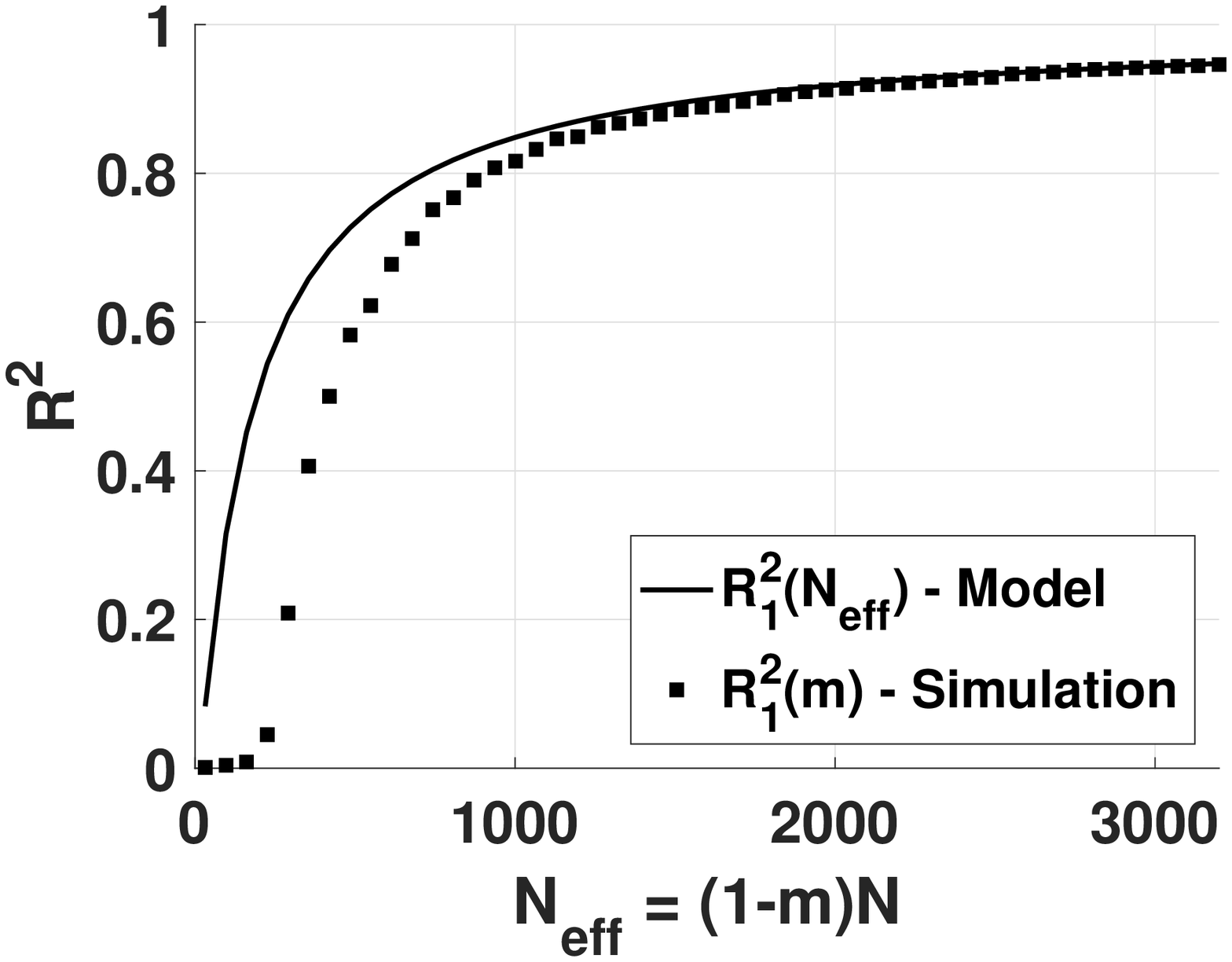}
        \caption{ }
        \label{fig:three}
    \end{subfigure}
        \begin{subfigure}[b]{0.45\linewidth}
        \includegraphics[width=\textwidth]{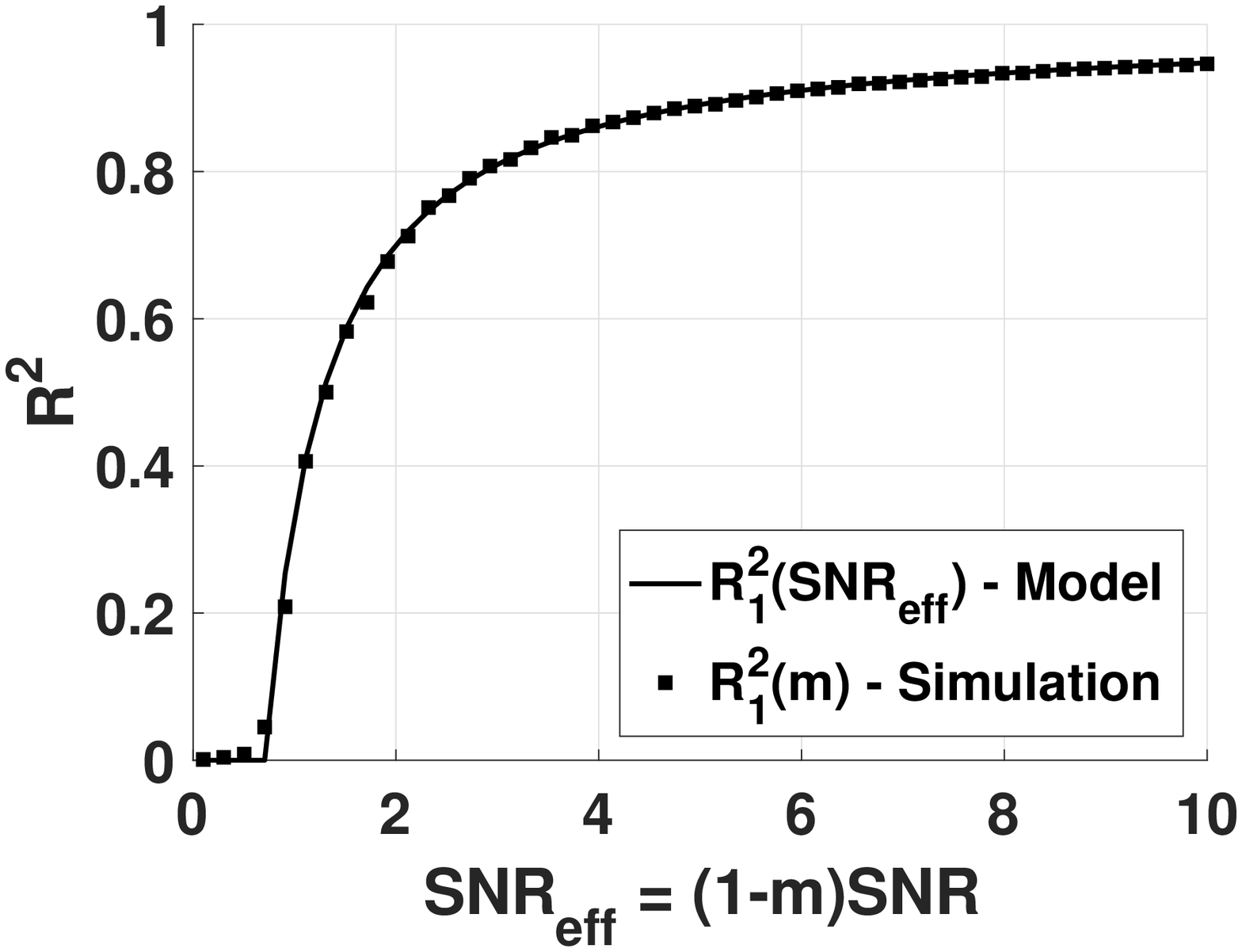}
        \caption{ }
        \label{fig:four}
    \end{subfigure}
    \caption{\\
    (a) Learning curves for the first two principal components, varied missing rate.
    (b) Learning curves for the first principal component at three different missing rates, varied signal-to-noise ratio.
    (c) Learning curve for the first principal component, varied missing rate. The full line represents the learning curve that would result from an effective sample size as $N(m) = (1-m)N$ in Eq.\ (\ref{eq:one}).
    (d) Learning curve for the first principal component, varied missing rate. The full line represents the learning curve that would result from an effective signal-to-noise ratio as $S(m) = (1- m)S $, i.e. the proposed relation in Eq.\ (\ref{eq:two}).
    }
    \label{fig:onetwothreefour}
    \end{center}
\end{figure}

We have applied probabilistic PCA to high dimensional datasets and simulated data, to investigate how missing data influences the principal components learned. As illustrated in Figure \ref{fig:fig1} an increasing missing rate implies a decrease in how well we can expect signal structures to be inferred using probabilistic PCA. A phase transition occurs in the learning curves at a critical missing rate, similarly to what is known to happen in the complete data case.
The theoretical result Eq.\ (\ref{eq:two}) is remarkably accurate as seen in Figure \ref{fig:one} and \ref{fig:two}.
\section{Design of simulation studies and relevance for real world data}\label{sec:methods}
We generate learning curves on simulated datasets and five high dimensional datasets: Faces in the Wild \cite{berg2005s}, MNIST \cite{lecun1998gradient}, NCI60 Cancer Microarray Project \cite{ross2000systematic}, a food pairing dataset \cite{ahn2011flavor} and FashionMNIST \cite{xiao2017/online}.

The generative process in probabilistic PCA assumes that data arise as a linear mapping of a $k$-dimensional, normally distributed latent variable $\z$ into the $D$-dimensional data space $\x$ through the transformation matrix $\Ab$, $ \x = \Ab\z + \epsilo, \quad \z \sim \mathcal{N}(\0,\I), \quad \epsilo \sim \mathcal{N}(\0,\sigma^2\I)$.
The signal is degraded by isotropic zero mean noise of variance $\sigma^2$. By assigning values to $\Ab$ and $\sigma^2$ we implicitly control the signal-to-noise ratio, the ratio between the signal variance and noise variance $S_i = \frac{||\ab_i||^2}{\sigma^2}$, where $\ab_i$ is the $i$'th column of the $\Ab$ matrix, i.e., the $i$'th signal direction. When a full dataset has been generated a missing data process is introduced, for this purpose missing completely at random. At a given missing rate $m$, each element of the data matrix is set to `missing' with probability $m$.

For a range of missing rates the signal directions can be estimated using PPCA and subsequently compared to the true symmetry breaking directions used to create the data, in terms of the squared cosine similarity, $R^2$. This experiment can be repeated a number of times to produce an estimate of the expectation $\langle R^2 \rangle_{\mathcal{X}}$.
To show the dependency on the signal-to-noise ratio, simulations have also been made where the noise $\sigma^2$ is varied instead of the missing rate. For each repetition a new noise-free dataset is generated and data-matrix elements are set to 'missing' according to a fixed missing rate. For a range of noise variances, noise is added to the data and signal directions are estimated using PPCA, to get an estimate of the expectation $\langle R^2 \rangle_{\mathcal{X}}$.

In order to apply the theory to the real datasets, we estimate and manipulate their signal-to-noise ratios. The signal-to-noise ratio can be estimated from the eigenvalues of a singular value decomposition. First the noise variance is estimated as the average of the eigenvalues in the non-signal dimensions. As the noise is isotropic, the variance in the signal direction consists of the signal variance plus the noise variance. The signal variance is therefore obtained as the eigenvalue in the signal direction minus the noise variance. Finally the signal-to-noise ratio is the ratio between the signal variance and the noise variance $\hat{\sigma}^2 = \frac{1}{D - k} \sum_{i=k+1}^D \lambda_i$,
$S_i = \frac{\lambda_i - \hat{\sigma}^2}{\hat{\sigma}^2}$,
where $k$ is the number of signal directions.
In order to manipulate the signal-to-noise ratio, zero-mean Gaussian noise with isotropic variance $\sigma_{\text{added}}^2$ is added to the dataset and the resulting signal-to-noise ratio becomes
$S_i = \frac{\lambda_i - \hat{\sigma}^2}{\hat{\sigma}^2 + \sigma_{\text{added}}^2}$

\section{Results}
Figure \ref{fig:one} was generated using 50 missing rates, $N=2000$, $D=3000$ and $\sigma^2=0.05$. Furthermore we set $||\ab_1||, ||\ab_2|| = 1.0, 0.5$ resulting in signal-to-noise ratios of $S_1 = 20$ and $S_2 = 5$. The full and dashed lines represent our proposed theory, Eq.\ (\ref{eq:two}), and the markers are simulation results averaged over 5 repetitions, approximating the expectation $\langle R^2\rangle_{\mathcal{X}}$. \\
Following the learning curves from right to left; when everything is missing nothing can be learned about the underlying signal direction. As we decrease the missing rate  a critical point is reached where learning for the 1st signal direction starts, with a steep phase transition. As even lesser data is missing, learning of the 2nd signal direction also initiates with a sharp phase transition.  After these steep increases, the learning curves enter a more flat plateau phase where learning is robust to the missing rate.

In figure \ref{fig:two} we vary the signal-to-noise ratio for three different fixed missing rates, all three learning curves are for the 1st signal direction. Here we see that with higher missing rates, we need a better signal-to-noise ratio before learning starts for the 1st signal direction. \\
The figure was generated using 100 signal-to-noise ratios, $N=3200$ and $D=1600$. The signal-to-noise ratio was obtained by adding zero-mean gaussian noise of different magnitude in variance to the linear transformation, keeping $||\ab_1|| = 1$. We plot averages over 10 datasets (error bars are smaller than the marker size).

Following \cite{schafer2002missing}, we could hypothesize that missing data changes the learning curves by decreasing the effective number of samples in the data, in the simplest case this would amount to $N(m) = (1 - m)N$. However, the theory in Eq.\ (\ref{eq:two}) suggests that the missing rate affects the learning curves by reducing the effective signal-to-noise ratio as $S(m) = (1-m)S$. This is illustrated in figure \ref{fig:three} and \ref{fig:four}.
Figure \ref{fig:three} compares the theory in Eq.\ (\ref{eq:one}) assuming an effective sample size to corresponding missing rate simulations. In figure \ref{fig:four} the theory in Eq.\ (\ref{eq:two}) is compared to corresponding missing rate simulations.

\begin{figure}[ht]
\vskip 0.2in
    \centering
    \begin{subfigure}[b]{0.45\linewidth}
        \includegraphics[width=\textwidth]{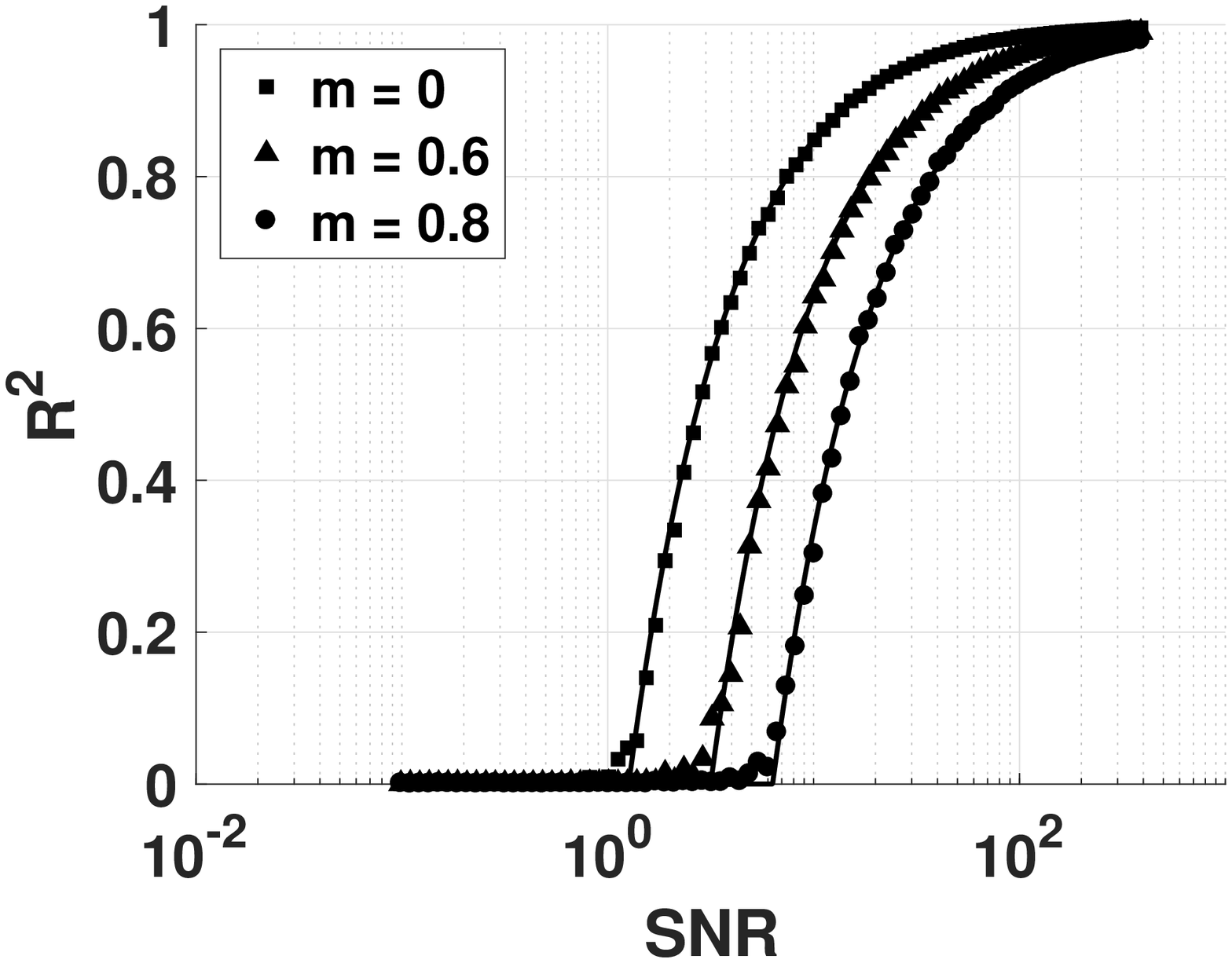}
        \caption{Faces in the Wild}
        \label{fig:wildfaces}
    \end{subfigure}
    \begin{subfigure}[b]{0.45\linewidth}
        \includegraphics[width=\textwidth]{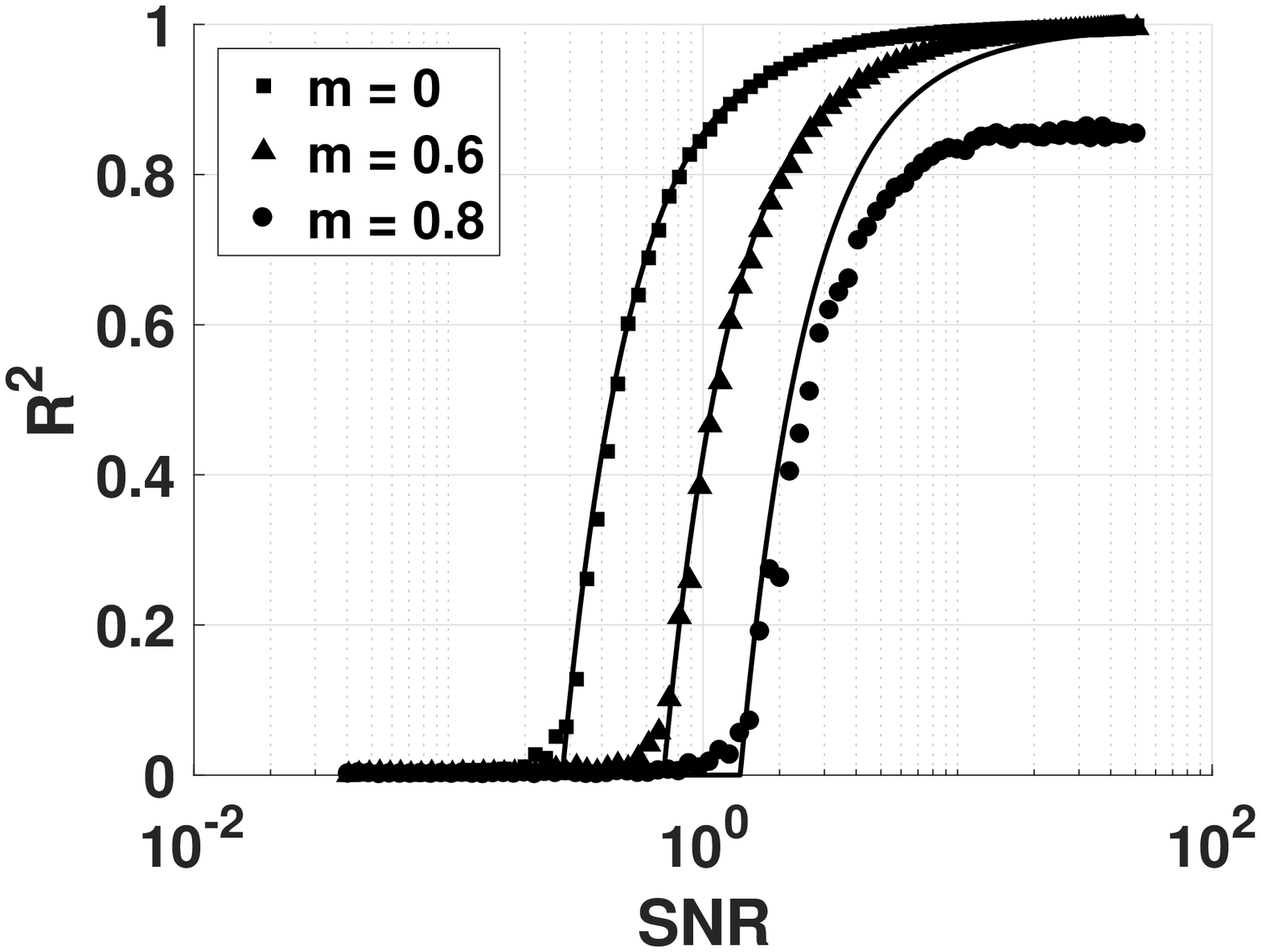}
        \caption{MNIST}
        \label{fig:mnist}
    \end{subfigure}
    \begin{subfigure}[b]{0.45\linewidth}
        \includegraphics[width=\textwidth]{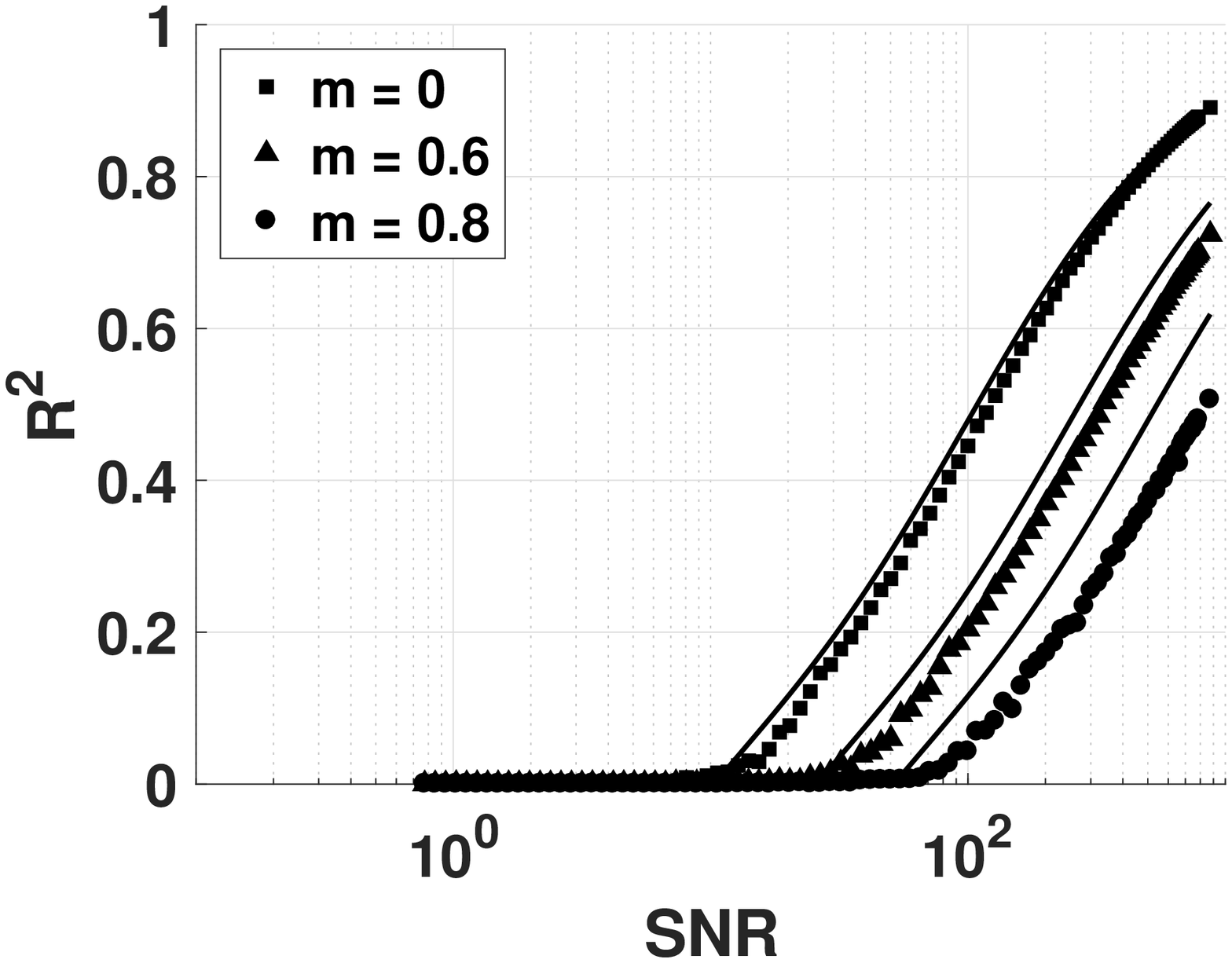}
        \caption{NCI60}
        \label{fig:micro}
    \end{subfigure}
    \begin{subfigure}[b]{0.45\linewidth}
        \includegraphics[width=\textwidth]{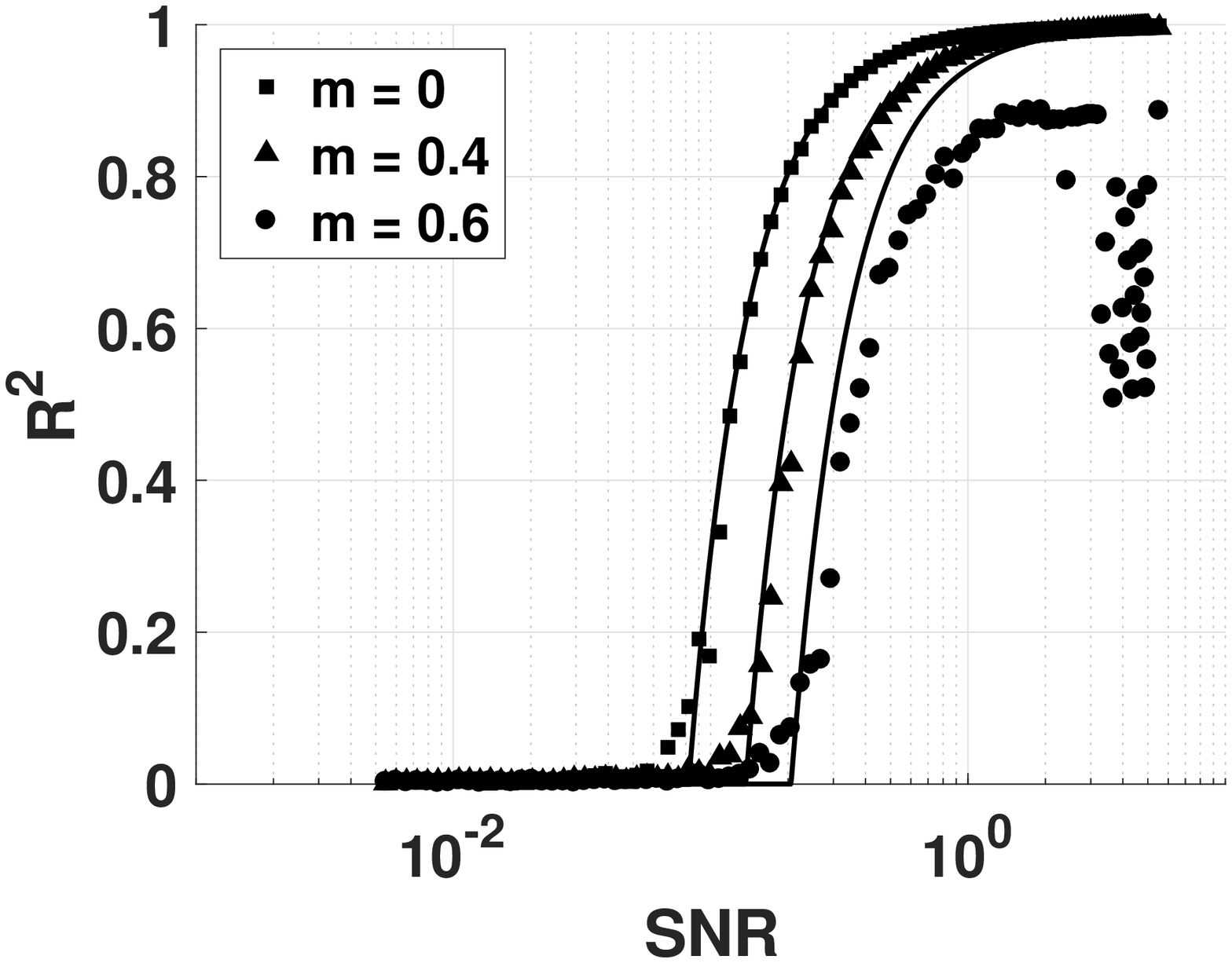}
        \caption{Food Pairing}
        \label{fig:food}
    \end{subfigure}
    \begin{subfigure}[b]{0.45\linewidth}
        \includegraphics[width=\textwidth]{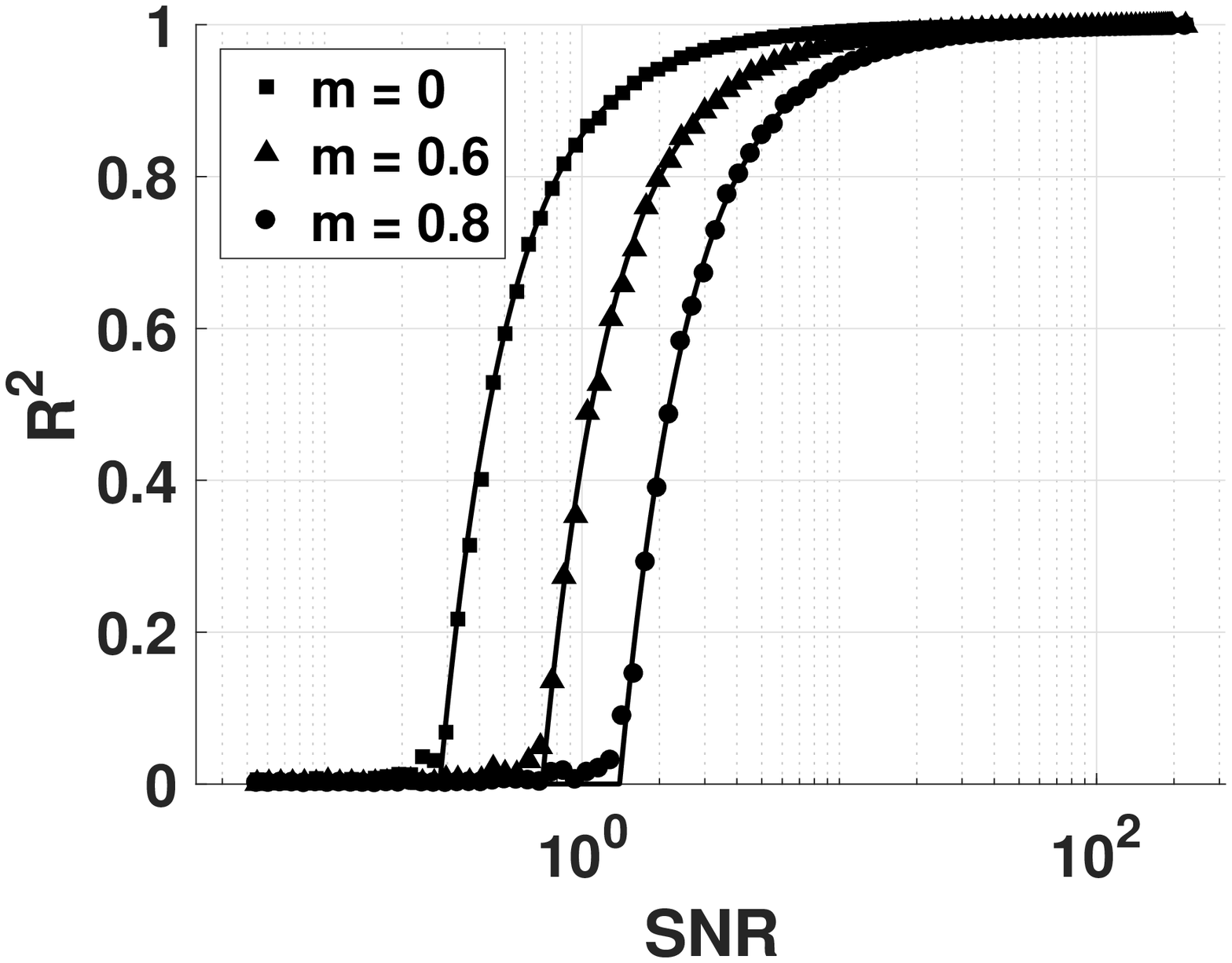}
        \caption{FashionMNIST}
        \label{fig:fashion}
    \end{subfigure}
    \caption{Learning curves for varying signal-to-noise ratio at different missing rates for five different high dimensional datasets. Plot symbols are averages over 10 different noise simulations and the full lines are learning curves as described by Eq.\ (\ref{eq:two}).}
    \label{fig:realdata}
\end{figure}

In five high dimensional datasets we manipulate the signal-to-noise ratio and missing rate in order to investigate the learning curves for the 1st  signal direction. Results are found in figure \ref{fig:realdata}, three learning curves for each dataset, each of them for the first signal direction at a fixed missing rate for varying signal-to-noise ratio.

\textbf{Faces in the Wild}\\
This is a collection of face images  captured under natural conditions and the principal components are eigenfaces capturing natural face appearance variation \cite{turk1991eigenfaces}. The first signal direction in the dataset is found using regular PCA in the no-missing case and used as a surrogate for the true signal direction. Now we set a fixed missing rate and the signal-to-noise ratio is gradually decreased by adding gaussian noise of increasing magnitude, to get the learning curves seen in figure \ref{fig:wildfaces}. Our theory, Eq.\ (\ref{eq:two}) is seen to match this dataset very well.\\
Plot symbols are averages over 10 datasets and solid lines are theoretical learning curves as described by Eq.\ (\ref{eq:two}). The estimate of the true signal direction deteriorates slowly as more and more noise is introduced, until a phase transition occurs and learning stops. At higher missing rates learning stops at a higher signal-to-noise ratio, i.e. missing data affects learning so that we need a better signal-to-noise ratio in order to learn with the same learning performance as in the non-missing case.

\textbf{MNIST}\\
The corresponding analysis for the handwritten digit dataset is provided in figure \ref{fig:mnist}. For a missing rate less than 0.6 the learning curve quite accurately follows the theory, while at larger rates, there are some deviations between theory and experiment. We observe a pronounced phase transition and that an increased signal-to-noise ratio is needed for higher missing rates before learning starts. Even though the MNIST dataset and the Faces in the Wild dataset are both image databases, the differences in the general appearance of the images between the two databases are affecting compliance with the PPCA model assumptions. Probabilistic PCA assumes an isotropic noise variance and this assumption is violated in the MNIST dataset, where there are pixel locations along the image boundary where there is no variation at all throughout the dataset.

\textbf{NCI60}\\
The microarray dataset is a collection of expression levels of genes in a number of different cell types. The signal variance in this dataset is due to different genes having the same kind of expression across cell types and the principal components are referred to as eigengenes \cite{alter2000singular}. Learning curves for the micro array dataset are shown in figure \ref{fig:micro}. For a missing rate of 0.6 and 0.8 an increasing deviation between the theoretical learning curve and the actual learning curve is seen. This dataset only has $N=64$ samples and therefore we may see finite size effects.

\textbf{Food pairing}\\
The food pairing dataset consists of food recipes, each of which containing one or more ingredients. The learning curves are found in figure \ref{fig:food}. The model assumptions for PPCA are violated in the sense that we are not looking at continuous data, but a binary indication of whether an ingredient was present in a recipe or not. Some ingredients are only in one recipe, leaving very little variance in this direction ($\sim 1/N = 0.000017$). This may explain the poor adherence for this dataset to the model curves. We have changed the missing rates here from 0.6 and 0.8 to 0.4 and 0.6 instead. The learning curve at a missing rate of 0.8 does not learn.

\textbf{FashionMNIST}\\
In figure \ref{fig:fashion} we present learning curves for the FashionMNIST dataset, which is designed to match the MNIST data in dimensions, sample size etc, but different in image contents. Here we have enough variation in all feature dimensions that this dataset adheres to the generative model of PPCA, and the learning curves accurately follow the model.

\begin{figure}[h]
  \centering
  \vspace{60pt}
  \includegraphics[width=\linewidth]{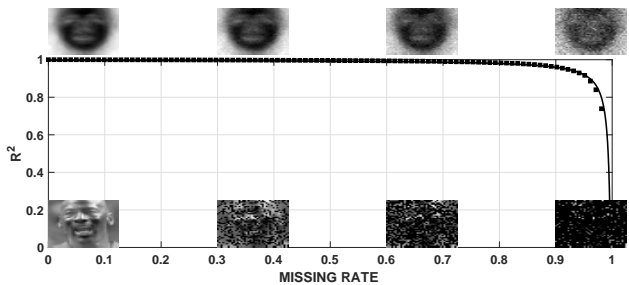}
  \caption{Missing rate experiment for the Faces in the Wild dataset. The images in the bottom row are example images illustrating the missing rate. The images in the top row are the corresponding first principal component as found by PPCA.}
  \label{fig:wild_fig1}
\end{figure}

Overall, our theory fits simulation studies while the theory seems to match real world datasets with deviations. The Faces in the Wild dataset adheres very well to the theory and a corresponding $\langle R^2\rangle_{\mathcal{X}}$ experiment where only the missing rate is varied can be seen in figure \ref{fig:wild_fig1}. On the other hand the MNIST and Food Pairing datasets show some clear deviations between experimental and theoretical learning curves. These deviations may be due to violations of the PPCA model assumption that the noise variance is isotropic. This assumption is definitely violated in the MNIST dataset as there are several dimensions with no variance at all, for example corner pixels, and the covariance matrix becomes degenerate. Similar observations are made in the food pairing dataset. The PPCA model is limited in its noise variance modelling and a more flexible model like factor analysis could potentially better capture the noise in these data.

\section{Discussion}
We have discussed the effects of missing data on our ability to learn signal structure.
We generalized  the well-established theoretical results for  principal components (Biehl and Mietzner, 1993) to include missing data, to find a result which is surprising in two ways. First, the generalization is remarkably simple modification of the known result for learning in PCA. Secondly, we find that the effect of missing data scales the effective signal-to-noise ratio, rather than scaling the sample size which has been earlier thought. 

The theory further predicts a phase transition in the learning curves as the missing rate increases. We used probabilistic principal component analysis for estimating signal structures in datasets with missing data and found consistent results in both in simulation data and in real datasets. Hence in conlusion: Missing data reduces signal-to-noise and give rise to a phase transition as the missing rate increases - from almost perfect learning to no learning at all.

\section*{Appendix A: Learning curves in the limit of large data}
In the simple PCA model we seek a single direction $\ab \in \mathbb{R}^d $ for which the mean projection is maximal
\begin{equation}
\widehat{\ab} = {\arg \max}_{||\vb|| = 1 } \sum_{n=1}^N (\vb^{T}\cdot\xb_n)^2.
\end{equation}
The key question is  how $\widehat{\ab}(\mathcal{X})$ relates to $\ab_0$ as a function of $N,D,\sigma^2$?
This problem has been the subject of quite detailed analysis in the machine learning / statistical mechanics community  \citep{hansel1992memorization,biehl1993statistical,watkin1993statistical,reimann1996learning,buhot1998phase}.
A particularly detailed account generalizing the transition to more general PCA was given in \cite{hoyle2007statistical} using the replica approach.
The first use of the replica method for such unsupervised learning is often attributed to the Biehl and Mietzner preprint of the paper \cite{biehl1993statistical}.
Earlier work on closely related supervised models include \cite{sompolinsky1990learning,seung1992statistical,biehl1993statistical}.

We are interested in quantifying the typical values of $(\ab_0^{T}\cdot\ab)^2$ as function of the physical parameters, hence we follow  \cite{hoyle2007statistical} and define the generating function (partition function),
\begin{equation}
Z(\hb)  =  \int d\ab\delta(||\ab||^2 -1) \exp\left( \beta\sum_{n=1}^N(\ab^{T}\cdot\xb_n)^2 + \hb^T\cdot\ab\right).
\end{equation}
The quantity of interest for a specific data set can be obtained directly from $Z$
\begin{equation}
 R^2 = (\ab_0^T\cdot\ab(\mathcal{X}))^2 = \lim_{\hb \rightarrow 0, \beta \rightarrow \infty} \left( \frac{\partial \log Z(\hb) }{\partial \hb}^T\cdot\ab_0\right)^2.
\end{equation}
It is not hard to see that
\begin{equation}
 (\ab_0^T\cdot\ab(\mathcal{X}))^2 = (\ab_0^T\cdot\ub_1(\mathcal{X}))^2,
\end{equation}
where $\ub_1(\mathcal{X})$ is eigenvector with the maximal eigenvalue of the empirical covariance matrix $\widehat{\Cb} = \frac{1}{N}\sum_{n=1}^N\xb_n\xb_n^{T}$.

More interesting is it to find the typical value of $R^2$, defined as
\begin{equation}
 \langle (\ab_0^T\cdot\ab(\mathcal{X}))^2 \rangle_{\mathcal{X}}= \int (\ab_0^T\cdot\ub_1(\mathcal{X}))^2 \prod_{n=1}^N p(\xb_n|\Cb) d\xb_n,
\end{equation}
where the data follows a zero mean multivariate normal law $p(\xb|\Cb)$ with covariance matrix $\Cb = \sigma^2 \Ib + \ab_0\ab_0^T$.
Unfortunately, it is entirely non-trivial to compute the distribution of this eigenvector.
A brute force calculation entails the averaging of the log-characteristic function with respect to the data distribution. In statistical mechanics this is approached using a series of tricks referred to as "the replica method". For completeness we here first reproduce the arguments for PCA and then make a generalization to the case with missing data. The first step is to invoke the limit
\begin{equation}
 \log Z  = \lim_{w \rightarrow 0} \frac{Z^w -1}{w}
\end{equation}
and use the fact that averaging powers of $Z$ is more feasible than averaging the logarithm.
Formally, the $w$'th power of $Z$ can be written
\begin{align*}
Z^w(\mathcal{X})  = & \int \prod_{j=1}^w d\ab_j\delta(||\ab_j||^2 -1) \\
& \exp\left( \beta\sum_{n=1}^N\sum_{j=1}^w(\ab_j^{T}\cdot\xb_n)^2 + \sum_{j=1}^w\hb^T\cdot\ab_j\right).
\end{align*}
The average over $\mathcal{X}$ then becomes
\begin{align*}
  \langle Z^w(\mathcal{X})\rangle_{\mathcal{X}}  = &\int \prod_{n=1}^Np(\xb_n|\Cb)d\xb_n \int \prod_{j=1}^w d\ab_j\delta(||\ab_j||^2 -1) \\
  & \exp\left( \beta\sum_{n=1}^N\sum_{j=1}^w(\ab_j^{T}\cdot\xb_n)^2 + \sum_{j=1}^w\hb^T\cdot\ab_j\right).
\end{align*}
When exchanging the sequence of integrals the average appears as an integral over a product of $N$ identical terms, i.e.,
\begin{align*}
  & \langle Z^w(\mathcal{X})\rangle_{\mathcal{X}}  = \int \prod_{j=1}^w d\ab_j\delta(||\ab_j||^2 -1) \\
  & \left( \int p(\xb|\Cb)d\xb \exp\left( \beta\sum_{j=1}^w(\ab_j^{T}\cdot\xb)^2 + \sum_{j=1}^w\hb^T\cdot\ab_j\right)\right)^N.
\end{align*}
For simplicity, we ignore the generating function variable $\hb$ and rewrite the quadratic terms
\begin{align*}
    &\exp\left( \beta(\ab_j^{T}\cdot\xb)^2\right) =  \exp\left( \frac{1}{2}2\beta(\ab_j^{T}\cdot\xb)^2\right) \\
    & =\int\prod_{j=1}^w  \frac{du_j}{\sqrt{2\pi}}\exp\left(-\frac{1}{2}u_{j}^2 + \sqrt{2\beta}u_{j}\ab_j^{T}\cdot\xb_n \right),
\end{align*}
this introduces a $w$ dimensional array of normal integrals over replica parameters $u_j$. We next compute the data integral
\begin{eqnarray}
 && \int p(\xb|\Cb)d\xb  \exp\left( \beta\sum_{j=1}^w(\ab_j^{T}\cdot\xb)^2 \right) \nonumber \\
 &=&  \int p(\xb|\Cb)d\xb \int\prod_{j=1}^w \frac{du_j }{\sqrt{2\pi}}\exp\left(-\frac{1}{2}\sum_{j=1}^w u_{j}^2 + \sum_{j=1}^w \sqrt{2\beta}u_{j}\ab_j^{T}\cdot\xb_n \right) \nonumber \\
&=&  \int\prod_{j=1}^w \frac{du_j}{\sqrt{2\pi}}\exp\left(-\frac{1}{2}\sum_{j=1}^w u_{j}^2 \right) \nonumber \\
&& \int p(\xb|\Cb)d\xb  \exp\left( \sum_{j=1}^w \sqrt{2\beta}u_{j}\ab_j^{T}\cdot\xb_n \right).
\label{int14}
\end{eqnarray}
The Gaussian data integral can be performed using
\begin{align}
& \int p(\xb|\Cb)d\xb  \exp\left( \sum_{j=1}^w \sqrt{2\beta}u_{j}\ab_j^{T}\cdot\xb \right) \nonumber \\
& =\exp\left( \beta \sum_{j,j'=1}^w u_{j}u_{j'}\ab_j^{T}\Cb^{-1}\ab_{j'} \right).
\label{int15}
\end{align}
Combining Eqs.\  (\ref{int14}) and (\ref{int15}) we get the final replica integral
\begin{align*}
 & \langle Z^w(\mathcal{X})\rangle_{\mathcal{X}} =  \int \prod_{j=1}^w d\ab_j\delta(||\ab_j||^2 -1) \\
 & \left(\int\prod_{j=1}^w \frac{du_j}{\sqrt{2\pi}}\exp\left(-\frac{1}{2}\sum_{j=1}^w u_{j}^2 + \beta \sum_{j,j'=1}^w u_{j}u_{j'}\ab_j^{T}\Cb\ab_{j'} \right)\right)^N.
\end{align*}
To proceed, we use $\Cb = \sigma^2 \Ib +\ab_0\ab_0^T$, to re-write the Gaussian integral over replica variables
\begin{align}
 & \langle Z^w(\mathcal{X})\rangle_{\mathcal{X}} =  \int \prod_{j=1}^w d\ab_j\delta(||\ab_j||^2 -1) \nonumber \\
 & \left(\int\prod_{j=1}^w \frac{du_j}{\sqrt{2\pi}}\exp\left(-\frac{1}{2}\sum_{j,j'=1}^w
 u_jM_{j,j'}u_{j'}\right)\right)^N
 \label{Eq19}
\end{align}
with
\begin{equation}
M_{j,j'}=\delta_{j,j'} - 2\beta\sigma^2\left(q_{j,j'} + S r_j r_{j'}\right),
\end{equation}
and where we have defined the key quantities $q_{j,j'}= \ab_j^{T}\cdot\ab_{j'}$ and $r_j = \ab_j^{T}\cdot\ab_{0}$ relating the inferred and generating parameters in the replicated system.
The Gaussian integrals yield
\begin{align*}
  &\langle Z^w(\mathcal{X})\rangle_{\mathcal{X}}  = \\
  & (2\pi)^{-Nw}\int \prod_{j=1}^w d\ab_j\delta(||\ab_j||^2 -1)|\Mb(\{\ab_j\})|^{-N/2}.
\end{align*}
The integrals over the $w$ unit spheres are next parameterized as integrals over the $q$ matrix and the $r$ vector,
\begin{eqnarray}
 && \langle Z^w(\mathcal{X})\rangle_{\mathcal{X}} = (2\pi)^{-Nw}\int \prod_{j=1}^w d\ab_j\delta(||\ab_j||^2 -1) \times \nonumber \\
 && \int\prod_{j>j'}dq_{j,j'}\delta(q_{j,j'}-\ab_j^T\cdot\ab_{j'}) \nonumber \\
 && \int\prod_{j}dr_{j}\delta(r_{j}-\ab_j^T\cdot\ab_{0}) |\Mb(\Qb,\rb)|^{-N/2}.
\end{eqnarray}
As the delta functions relate $\Qb$ and the inner products of $\ab$, we can replace $\delta(||\ab_j||^2 -1)$ by $\delta(q_{j,j} - 1)$.
The $\Qb$ and $\rb$ integrals can be combined by defining the matrix
 \begin{equation}
 \Qb_{+}=
\begin{bmatrix}
  \ab_0^T\cdot \ab_0 & \ab_0^T\cdot \ab_1 &...  & \ab_0^T\cdot \ab_w  \\
  \ab_1^T\cdot \ab_0 & \ab_0^T\cdot \ab_1 &...  & \ab_1^T\cdot \ab_w  \\
  \vdots & \vdots &\ddots  & \vdots  \\
  \ab_w^T\cdot \ab_0 & \ab_w^T\cdot \ab_1 &...  & \ab_w^T\cdot \ab_w  \\
  \end{bmatrix} =
  \begin{bmatrix}
    1 & \rb \\
    \rb^T & \Qb \\
  \end{bmatrix}.
\end{equation}
With this definition,
\begin{eqnarray}
 && \langle Z^w(\mathcal{X})\rangle_{\mathcal{X}} = (2\pi)^{-Nw}\int \prod_{j=1}^w d\ab_j \times \nonumber \\
 && \int d\Qb_{+}\prod_{j>j'=0}^w\delta(\Qb_{+,j,j'}-\ab_j^T\cdot\ab_{j'}) \nonumber \\
 && \prod_{j=0}^w\delta(\Qb_{+,j,j}-1) |\Mb(\Qb,\rb)|^{-N/2},
\end{eqnarray}
the $\ab$ integrals can be performed to leading order in $D$ with a procedure outlined in Appendix A of  \cite{ahr1999statistical}. 
First we write the integral in matrix form with $\Ab$ a $D \times (w+1)$ matrix
\begin{equation}
\int  d\Ab \delta(\Qb_{+}-\Ab^T\cdot\Ab) = |\Ub^T\Lambdab|^D \int d\tilde{\Ab} \delta(\Ib-\tilde{\Ab}^T\cdot\tilde{\Ab}).
\end{equation}
Here we used the transformation $\Ab \rightarrow \tilde{\Ab} = \Ab\Ub\Lambdab$ based on
 the spectral representation of $(w+1)\times (w+1)$ matrix  $\Qb_{+} = \Ub^T\Lambdab\Lambdab\Ub$. Here $\Ub$ is an orthogonal matrix and $\Lambdab$ is a diagonal matrix with the square roots of the eigenvalues of symmetric real matrix $\Qb_{+}$. The integral $\int d\tilde{\Ab} \delta(\Ib-\tilde{\Ab}^T\cdot\tilde{\Ab})$ is seen to be a constant with respect to the integration variables $\Qb_{+}$. Hence, to leading order the dependent term is
 \begin{equation}
 \int  d\Ab \delta(\Qb_{+}-\Ab^T\cdot\Ab) \propto |\Ub^T\Lambdab|^D \propto |\Qb_{+}|^{\frac{D}{2}}.
 \end{equation}
 The determinant can be computed using the identity
 \begin{equation}
\left|\begin{matrix}
  \Xb_{11} &  \Xb_{12}   \\
    \Xb_{21} &  \Xb_{22}   \\
  \end{matrix}
\right| = |\Xb_{11}||\Xb_{22}-\Xb_{21}\Xb_{11}^{-1}\Xb_{12}  |
\end{equation}
with $\Xb_{11} = 1$, $\Xb_{21} = \rb$ and $\Xb_{22} = \Qb$,
%
\begin{equation}
 \langle Z^w(\mathcal{X})\rangle_{\mathcal{X}} \propto  \int\prod_{j>j'}dq_{j,j'}\int\prod_{j}dr_{j}|\Qb -\rb\rb^T |^{D/2}\Mb(\Qb,\rb)|^{-N/2}.
\end{equation}
To bring the scaling with $D,N \rightarrow \infty$ out more clearly, we rewrite
\begin{eqnarray}
 && \langle Z^m(\mathcal{X})\rangle_{\mathcal{X}} \propto  \nonumber \\
 && \int d\Qb d\rb \exp\left[ D\left( \frac{1}{2}\log |\Qb -\rb\rb^T| \right. \right. \nonumber \\
 && \left. \left. - \frac{\alpha}{2}\log |\Ib - 2\beta\sigma^2(\Qb + S\rb\rb^T)|\right)\right],
\end{eqnarray}
with the definitions from the main text $\alpha = \frac{N}{D}, S = \frac{1}{\sigma^2}$. 

We are interested in the limit of large $D$, where the integral is dominated by the stationary points of the integrant. We further assume "replica symmetry", hence, the following simple structures of $\Qb, \rb$
\begin{equation}
\Qb =\begin{bmatrix}
  1 & q & ... & q  \\
  q & 1 & ... & q  \\
  \vdots & \vdots & \ddots & \vdots  \\
  q & q & ... & 1  \\
\end{bmatrix}, \ \
\rb = \begin{bmatrix}
         r \\
         r \\
         \vdots \\
         r \\
       \end{bmatrix}
\end{equation}
The two $w\times w$ determinants can then be computed using simple algebra
\begin{equation}
\left|
\begin{matrix}
  z & y & ... & y \\
  y & z & ... & y \\
  \vdots & \vdots & \ddots & \vdots \\
  y & y & ... & z \\
\end{matrix}
\right| = (1+ \frac{wy}{z-y})(z-y)^{w}.
\end{equation}
Hence,
\begin{equation}
 \log |\Qb -\rb\rb^T| = \log \left( 1 +\frac{w(q-r^2)}{1-q} \right) + w \log (1-q)
\end{equation}
\begin{align*}
& \log |\Ib - 2\beta\sigma^2(\Qb + S\rb\rb^T)|  \\
= & \log \left( 1 - w\frac{2\beta\sigma^2(q + S r^2)}{1-\beta\sigma^2(1- S r^2)}\right)
 + w \log \left( 1-2\beta\sigma^2(1-S r^2) \right).
\end{align*}
Finally, we send $w \rightarrow 0$ to get,
\begin{equation}
 \lim_{w \rightarrow 0}\frac{1}{w} \log |\Qb -\rb\rb^T| = \frac{q-r^2}{1-q},
\end{equation}
and
\begin{align*}
& \lim_{m \rightarrow 0}\frac{1}{w} \log |\Ib - \beta\sigma^2(\Qb + S\rb\rb^T)| \\
= & -\frac{2\beta\sigma^2(q + S r^2)}{1-2\beta\sigma^2(1- q)}.
\end{align*}
Following \cite{reimann1996learning} we locate the stationary points first wrt.\ $q$ for fixed $r$ and then find the resulting stationary point wrt.\ $r$.
To leading order the averaged log partition function is given by
\begin{equation}
\langle \log Z (\mathcal{X})\rangle_{\mathcal{X}} \propto  \frac{1+r^2}{v} + 2\alpha\frac{1 + S r^2}{1-2v},
\end{equation}
where $v=\beta\sigma^2(1-q)$. To simplify the location of the stationary points we further define the two functions of $r$: $\phi_1=1+r^2$ and $\phi_2=\alpha(1 + S r^2)$.
\begin{equation}
\langle \log Z (\mathcal{X})\rangle_{\mathcal{X}} \propto \left( \frac{\phi_1}{v} + \frac{\phi_2}{1-2v}\right).
\end{equation}
The stationary point is given by
\begin{equation}
\widehat{v} = \frac{1}{2}\frac{\sqrt{\phi_1}}{\sqrt{\phi_1} \pm \sqrt{\phi_2}}
\end{equation}
We choose the positive square root and substitute back into the log partition function to obtain
\begin{align*}
& \langle \log Z (\mathcal{X})\rangle_{\mathcal{X}} \propto  \left( \sqrt{\phi_1} \pm \sqrt{\phi_2} \right) \\
=& \left( \sqrt{1+r^2} + \sqrt{\alpha(1+S r^2)} \right)^2.
\end{align*}
Finally,  the extremal point wrt.\ $r$ is given by
\begin{equation}
\langle (\ab_0\cdot\ab)^2\rangle_{\mathcal{X}} = \widehat{r^2} = \begin{cases}
\frac{\alpha S^2 - 1}{S(1+\alpha S)} & \text{if $\alpha > \frac{1}{S^2}$} \\[3mm]    0 & \text{if $\alpha \le \frac{1}{S^2}$}
\end{cases}
\end{equation}
producing the non-smooth learning curve result, i.e., relation between $\ab$ and $\ab_0$ as a function of the relative sample size $\alpha$. It is remarkable that in the large scale system, there is a range of sample sizes $N < N_{\rm critical} \equiv D\sigma^4$ for which learning is completely absent.

\subsection*{Statistical mechanics of simple PCA with missing data}
To account for missing data we introduce the missing indicator $s_{n,d} \in [0,1]$, with $\sb_{n,d} = 1$ when the feature is present. The averaging procedure increases a bit in bookkeeping as we only average over the actual present features in each sample, hence, we obtain
\begin{align}
  & \langle Z^w(\mathcal{X})\rangle_{\mathcal{X}}  = \int_{||\ab_j||^2 = 1}d\Ab \prod_{n=1}^N\int p(\xb_n|\Cb_n)d\xb_n  \nonumber \\
  & \exp\left( \beta\sum_{n=1}^N\sum_{j=1}^w(\ab_j^{T}\cdot(\sb_n\odot\xb_n))^2 + \sum_{j=1}^w\hb^T\cdot\ab_j\right),
\label{Eq13missing}
\end{align}
where the element wise product is  $(\sb_n\odot\xb_n)_d = \sb_{n,d}\xb_{n,d}$, and $\Cb_n$ is the sub-matrix defined by the present features in the $n$'th sample.
Similarly, by performing the integration of present features, Equation \ref{Eq19} becomes
\begin{align}
 & \langle Z^w(\mathcal{X})\rangle_{\mathcal{X}} =  \int_{||\ab_j||^2 = 1}d\Ab \prod_{n=1}^N
 \int\prod_{j=1}^w \frac{du_j^n}{\sqrt{2\pi}} \nonumber \\
 & \exp\left(-\frac{1}{2}\sum_{j,j'=1}^w
 u_j^n M_{j,j'}^n u_{j'}^n\right).
 \label{Eq19missing}
\end{align}
Here,
\begin{equation}
M_{j,j'}^n = \delta_{j,j'} - 2\beta\sigma^2\left(q_{j,j'}^n +S r_{j}^n r_{j'}^n\right),
\end{equation}
with $q_{j,j'}^n = \sum_{d=1}^D \sb_{n,d}\ab_{j,d}\ab_{j',d}$ and $r_{j}^n = \sum_{d=1}^D \sb_{n,d}\ab_{j,d}\ab_{0,j',d}$.

Re-writing the product over samples as an exponentiated sum over $M_{j,j'}^n$  terms, we may invoke a simple self-averaging assumption \cite{selfav/online} (the sum can be replaced by its average),
\begin{equation}
 \langle M_{j,j'}^n\rangle = \delta_{j,j'} - 2\beta\sigma^2\left((1-m) q_{j,j'}^n + (1-m)^2 S r_{j}^n r_{j'}^n\right),
\end{equation}
where $m$ is the missing rate and we used that the $q$ term involves a single $(1-m)$ factor while $r_{j}^n r_{j'}^n$ term involves two independent $(1-m)$ factors.

By inspection we find that this is equivalent to the earlier expression if $\beta \rightarrow (1-m)\beta$ and $S \rightarrow (1-m)S$.
Hence, the asymptotic expression for $R^2$ in presence of missing data at rate $m$ is as already shown in Equation \ref{eq:two},
\begin{equation}
\langle  R^2 \rangle_{\mathcal{X}} = 	%
    \begin{cases}\vspace*{-1mm}
          0	&	\alpha ((1-m)S)^{2} <1,  \\
           \frac{\alpha ((1-m)S)^2 - 1}{(1-m)S + \alpha ((1-m)S)^2}	&	\alpha((1-m)S)^{2} \geq 1.
	\end{cases}
\end{equation}

\newpage

\bibliography{lkh_nbip}
\bibliographystyle{icml2019}

\end{document}